\documentclass{article}
\usepackage{ijcai20}

\usepackage{times}
\usepackage{soul}
\usepackage{url}
\usepackage[hidelinks]{hyperref}
\usepackage{graphicx}
\usepackage{amsmath}
\usepackage{amsthm}
\usepackage{booktabs}
\usepackage{algorithm}
\usepackage{algorithmic}
\title{Automatic low-bit hybrid quantization of neural networks through meta learning}

\author{
Tao Wang$^1$\footnote{Contact Author}\and
Junsong Wang$^1$\and
Chang Xu$^1$\And
Chao Xue$^1$\\
\affiliations
$^1$IBM Research – China\\
\emails
\{wtaobjf, junsongwang, bjxchang, xuechao\}@cn.ibm.com
}

\begin{document}

\maketitle

\begin{abstract}
Model quantization is a widely used technique to compress and accelerate deep neural network (DNN) inference, especially when deploying to edge or IoT devices with limited computation capacity and power consumption budget. The uniform bit width quantization across all the layers is usually sub-optimal and the exploration of hybrid quantization for different layers is vital for efficient deep compression. In this paper, we employ the meta learning method to automatically realize low-bit hybrid quantization of neural networks. A MetaQuantNet, together with a Quantization function, are trained to generate the quantized weights for the target DNN. Then, we apply a genetic algorithm to search the best hybrid quantization policy that meets compression constraints. With the best searched quantization policy, we subsequently retrain or finetune to further improve the performance of the quantized target network.  Extensive experiments demonstrate the performance of searched hybrid quantization scheme surpass that of uniform bitwidth counterpart. Compared to the existing reinforcement learning (RL) based hybrid quantization search approach that relies on tedious explorations, our meta learning approach is more efficient and effective for any compression requirements since the MetaQuantNet only needs be trained once.
\end{abstract}

\section{Introduction}
Deep neural networks (DNN) are widely used for solving various artificial intelligent (AI) tasks, like image classification \cite{krizhevsky2012imagenet}, object detection \cite{girshick2014rich}, natural language processing (NLP) \cite{li2017deep} and deep reinforcement learning (RL) \cite{mnih2013playing}. However, the training, inference and storage of a modern deep neural network typically require powerful GPUs, dedicated hardware accelerators and storage resources, which hinders the wide applications of DNN to edge devices where memory and computational capacities are limited. Many research interests have been focused on compressing deep learning models without significant performance degradation to save computation cost and memory storage, such as pruning \cite{han2015learning,li2016pruning,liu2019metapruning}, quantization \cite{courbariaux2016binarized,choi2018pact,leng2018extremely} and knowledge distillation \cite{hinton2015distilling}.

For model quantization, many efforts have been made to reduce the model size and accelerate the model inference on various hardwares. It has been well demonstrated that direct quantizing the trained float-point model to 16 bits or 8 bits would not significantly degrade the accuracy. To achieve extremely higher energy efficiency in resource constrained edge devices, the extremely low bit quantization approach is proposed in literature \cite{courbariaux2015binaryconnect,li2016ternary,choi2018pact,rastegari2016xnor}, where use binary or ternary status to represent the weights and only use very limited bits to represent the activations, which can totally eliminate the multiplication operation. Even the full binary network XNOR \cite{rastegari2016xnor} is proposed to degenerated the computation to XNOR and pop-count.

However, most of the quantization approaches only investigate uniform bitwidth quantization across all layers of DNN, which is usually sub-optimal under certain compression constraints. The precision of each layer in a network has different influence to the final accuracy, which is discussed in \cite{wang2018design} and also varies with the architectures of the deep networks. The exploration of low-bit hybrid quantization of different network layers is vital for deep compression of DNN without accuracy degradation. The conventional hybrid quantization of DNN requires domain experts and some empirical rules to explore best hybrid quantization policy, which is used in \cite{wang2018design}. Recently, a RL based automatic hybrid quantization searching approach is also proposed in \cite{wang2019haq}. They demonstrate that the searched hybrid quantization can explore the compression ability of networks and outperform the uniform quantization. However, they only focus on the relatively high precision quantization, which can fine-tune from a float-point trained network to significantly reduce the search time. In extremely low-bit network case, it is impossible to directly fine-tune from a float-point trained network and tedious training process from scratch takes time and resources, which makes their RL based approach infeasible.

\begin{figure*}[htb]
\centering
\includegraphics[width=0.8\textwidth]{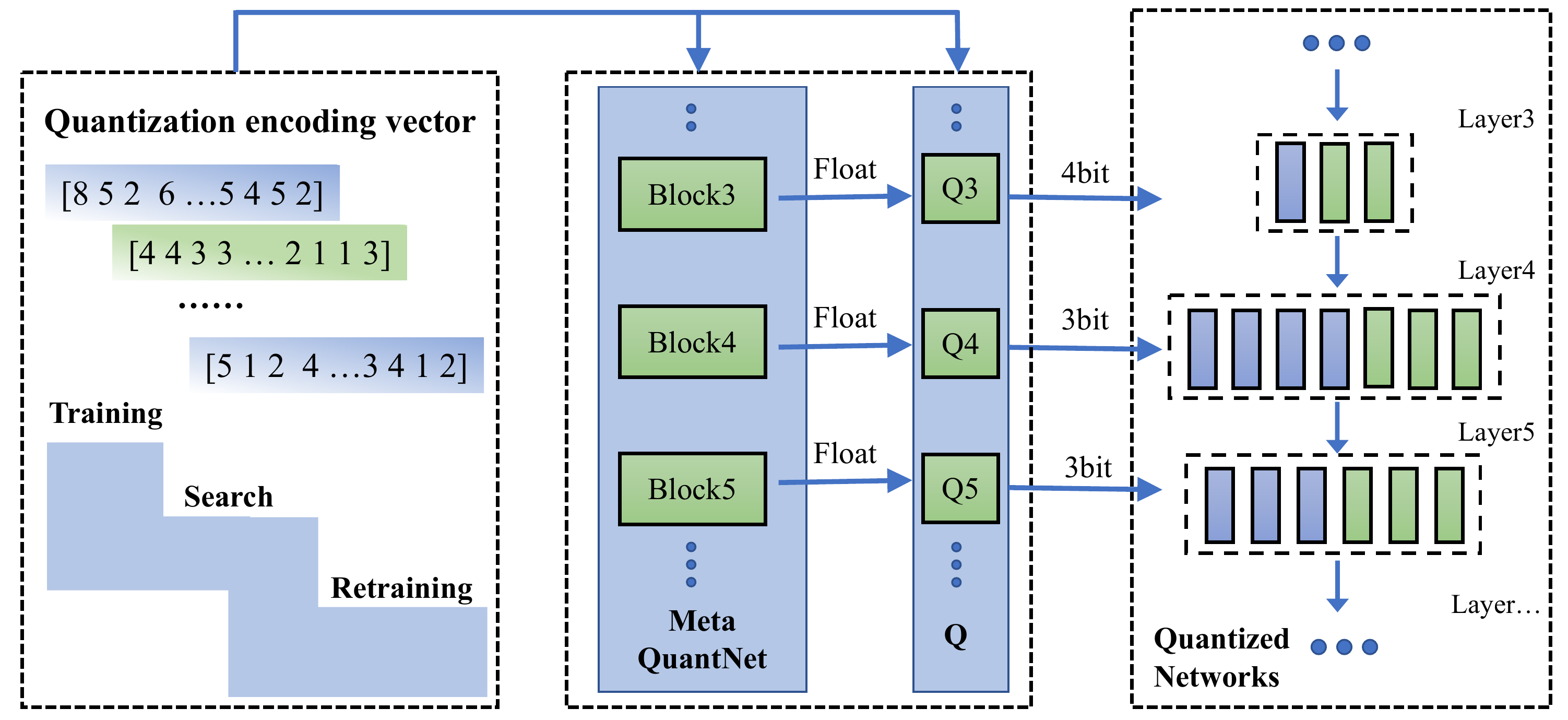}
\caption{An overview of our proposed automatic low-bit hybrid quantization of neural networks through meta learning. We utilize a MetaQuantNet as a hypernetwork to predict weights of each layer in the target quantized network. The predicted weights are quantized by a Quantization function (Q). The whole process is divided into three stages: 1) training the MetaQuantNet by input random generated hybrid quantization encoding vectors. Each value of the encoding vector represents the bitwidth of a corresponding layer and controls the of the Quantization function (Q) simultaneously. 2) search the best hybrid quantization encoding vector such that the target quantized network meets the model compression constraints but maintains the highest validation accuracy. 3) retraining the MetaQuantNet with the searched best hybrid quantization policy to further improve the accuracy of target quantized network. }
\label{fig:meta_quanta}
\end{figure*}

Our method is mainly inspired by the recent work of MetaPruning \cite{liu2019metapruning} and Hypernetworks \cite{ha2016hypernetworks}. The hypernetworks demonstrate a way to use an external network, known as a hypernetwork, to generate the weights for another network. In the work \cite{liu2019metapruning}, a hypernetwork PruningNet is built to generate weight parameters for various pruned target networks. The PruningNet is trained with inputs of random filter numbers as encoding vector for each layer. The optimal pruning structure of target network is obtained by searching the PruningNet. Similarly, we also adopt the meta learning framework to realize automatic low-bit hybrid quantization. Instead of seeking the best pruning structure, we propose to search the best hybrid quantization policy of a quantized network. In our method, We utilize a MetaQuantNet as a hypernetwork to predict weights of each layer in the target quantized network. The MetaQuantNet, together with a Quantization function (Q), are trained to generate the quantized weights for the target network. With certain constraints, the best hybrid quantization policy can be obtained by search the well trained MetaQuantNet. It is worth noticing that in this work we quantize all the layers of the target neural network and only focus on the weights quantization case.

The proposed method is illustrated in Figure \ref{fig:meta_quanta}. The whole process is divided into three stages: training, search and retraining. For the training process, stochastic sampling values between 1-8 that encode the quantization bitwidth are the inputs of the MetaQuantNet, and the quantization encoding vectors control the Quantization function (Q) simultaneously. Each number of the quantization encoding vector corresponds to the bitwidth of target quantized network layer. At second stage, we apply a genetic algorithm to search the best hybrid quantization combinations under certain constraints. Only the results of top-N performance that meet the constraints  can be preserved as parent genes to generate the off-springs. Finally, with the searched best hybrid quantization result, we continue to retrain or fine-tune MetaQuantNet to further improve the performance of the quantized target network. 

Compared to the existing reinforcement learning based hybrid quantization search approach like HAQ \cite{wang2019haq}, we find that the reinforcement learning search should be repeatedly executed if given various compression constraints. Each exploration process corresponds the same compression constraints. Best policy can not be obtained until the exploration finished. If the constraints change slightly, the exploration process should be repeated again. However, in our method the MetaQuantNet are trained with various hybrid quantization policy and thus acquire the meta-knowledge for these tasks. Once the MetaQuantNet is well trained, it can predict weights of target networks for various hybrid quantization encoding vector inputs. Hence the MetaQuantNet just needs to be trained once and best hybrid quantization policies can be fast searched for different compression constraints under the same workflow.

The primary contributions of this work include:
\begin{itemize}
    \item We propose a method for automatic low-bit hybrid quantization of neural networks through meta learning, which frees human efforts for designing hybrid bitwidth layer by layer. Besides, our approach can be easily combined with most existing AutoML techniques in an out-of-box fashion: after the optimal neural network structure is gained, a hybrid quantization can be used for further model compression.
    
    \item Compared to the existing RL based framework \cite{wang2019haq}, our method is more efficient and feasible in application. Once the MetaQuantNet is well trained, it can be applied under various compression requirements. Moreover, our method shows advantages in the abilities of realizing the extremely low-bit hybrid quantization. It is impossible to obtain the accuracy results for extremely low-bit quantized network by just finetuning float-point networks, which also makes their RL approach unfeasible.
    
    \item We show that the hybrid quantization strategy can always maintain higher accuracy than traditional uniform quantization policy in extremely low-bit quantization cases. DNN can be compressed more by adopting hybrid quantization policy without significant accuracy degradation.
    
    \item The searched best hybrid quantization policy can be various under different constraints. But we find that higher bitwidth is preferred for the first layer and last classification layer, which confirms the common design heuristics of hybrid quantization. Moreover, our experiments also show that there exists some layers always need much lower bitwidth representation for different kinds of tasks and constraints.
\end{itemize}

\section{Related Work}
\textbf{Quantization} Extensive research works have been carried out on low-bit quantization for model compression. \cite{han2015deep} proposed to use clustering method to push weights to quantized values. BinaryConnect and BinaryWeight networks have been proposed in \cite{courbariaux2015binaryconnect}. Besides, \cite{rastegari2016xnor} propose XNOR network to degenerate the computation to XNOR and pop-count operation. Those adopt the binary representation of network weights and activations, which compressed most of the networks but the accuracy drops significantly in their cases. \cite{li2016ternary} suggests to use ternary instead of binary and adopt a float scaling parameter to keep the performance. The quantized network training is a mainly problem for low bit quantization. Recently, \cite{choi2018pact} propose a novel technique named PACT to clip activations when quantizing both weights and activation during training. They achieve highest accuracy for both low bit weight and activations quantization. \cite{leng2018extremely} model the low bit network training as discretely constrained optimization problem and utilizes the ADMM method to decouple continuous parameters updates from discrete case. In these works, they do not consider the hybrid quantization strategy for target networks. Our method for hybrid quantization policy exploration can be combined with their training optimization techniques.

\noindent \textbf{AutoML and Meta Learning} 
AutoML have been widely studied to search neural network structures and hyper-parameters tailored to specific task and dataset with minimal human efforts. It has
achieved good successes in both vision and language. Existing AutoML works usually use methods based on genetic algorithms \cite{Real17}, random search \cite{Bergstra12}, Bayesian optimization \cite{Snoek12}, reinforcement learning \cite{Zoph17a} and continuous differentiable methods \cite{Liu18}. In our work, we adopt a genetic algorithm to explore a good hybrid quantization policy for the target network.
Our work adopt the same meta learning structure as the MetaPruning work \cite{liu2019metapruning}. Instead of search the pruning structure, we realize automatic low-bit hybrid quantization of target networks.

\section{Methodology}
In this section, we formulate our meta learning method for automatic low-bit hybrid quantization of neural networks under certain compression constraints.

The problem of search hybrid quantization policy of a neural network can be formulated as: 
\begin{equation}
\begin{array}{ll}
\mathop{\arg\min}\limits_{q_1,q_2,\dots,q_l}\mathcal{L}(\hat{Q}(q_1,q_2,\dots,q_l;W),X)  &  \\
s.t.\quad C< constraint, &
\end{array}
\label{eq:loss}
\end{equation}
where $\hat{Q}(q;W)$ is a quantization function that pushes neural network weights $W$ to nearby quantization levels. $X$ represents the input dataset. We only quantize weights $W$ and our goal is to find a best hybrid quantization policy $(q_1,q_2,\dots,q_l)$ for the deep neural network layers from $1^{st}$ to $l^{th}$ layers such that the loss $\mathcal{L}$ is minimum under constraints $C$. $q_l$ stands for quantization bitwidth of $l^{th}$ layer. The cost function $C$ is the target compression goal of certain constraints, such as the model size after quantization should be 10 times smaller than original float model size or the energy consumption should be reduced to a certain level with dedicated hardware accelerator.

\subsection{Quantization function}
We adopt the commonly used equally distributed quantization function that can be easily adapted to edge computing hardware \cite{choi2018pact}. The $q$ bitwidth quantization function $\hat{Q}(x)$ is defined as:
\begin{equation}
\hat{Q}(x)=\frac{1}{2^q-1}round((2^q-1)\cdot x)
\end{equation}

For back-propagation, the gradient of the quantization function $\hat{Q}(x)$ is approximated by Straight-Through Estimation (STE) method \cite{bengio2013estimating}:
\begin{equation}
\frac{\partial \hat{Q}(x)}{\partial x}=
\left\{
\begin{array}{lr}
1, \quad |x|<\Delta&  \\
0, \quad others &  
\end{array}
\right.
\end{equation}
where $\Delta$ is manually defined to bound gradients for larger input values.

In our method, a scaling function $sc$: $\mathcal{R} \rightarrow [0,1]$ is used to normalize arbitrary weights values to $[0,1]$ at first. The scaling function $sc(W)$ is defined as:
\begin{equation}
sc(W)=\frac{W-\beta}{\alpha}
\end{equation}
where $\alpha=W_{max}-W_{min}$ and $\beta=W_{min}$

Then weights are quantized like this:
\begin{equation}
\hat{W}=\alpha \cdot \hat{Q}(\frac{W-\beta}{\alpha}) +\beta
\end{equation}

The gradient of the loss function about weights are
\begin{equation}
\frac{\partial \mathcal{L}}{\partial W}=\frac{\partial \mathcal{L}}{\partial \hat{W}}\frac{\partial \hat{W}}{\partial \hat{Q}}\frac{\partial \hat{Q}}{W}=
\left\{
\begin{array}{lr}
\frac{\partial \mathcal{L}}{\partial \hat{W}}, \quad |W|<\Delta&  \\
0, \quad others & 
\end{array}
\right.
\end{equation}


\subsection{Hybrid quantization procedure}
In our method, we adopt the hypernetwork framework to generate weights of target network from a MetaQuantNet \cite{ha2016hypernetworks}. The MetaQuantNet takes the target network quantization encoding vector ($q_1,q_2,\dots, q_l$) as input and outputs weights for target networks:
\begin{equation}
W=MetaQuantNet(q_1,q_2,\dots, q_l;w)
\end{equation}
where $w$ is the float-point weights of MetaQuantNet that needed to be trained. $W$ are the generated weights for target network. To maintain accuracy of quantized network, people normally scale the quantized weights by with a scaling parameters $\gamma$, which can be obtained by minimizing $||W-\gamma \hat{Q}(W)||$ or directly estimating $\gamma$ with weight values like in paper \cite{li2016ternary,rastegari2016xnor}. However, we embed a branch structure inside the block in MetaQuantNet to predict the value $\gamma$ to realize an end-to-end training procedure, which is similar to \cite{leng2018extremely}.

\begin{figure}[htb]
\centering
\includegraphics[width=0.45\textwidth]{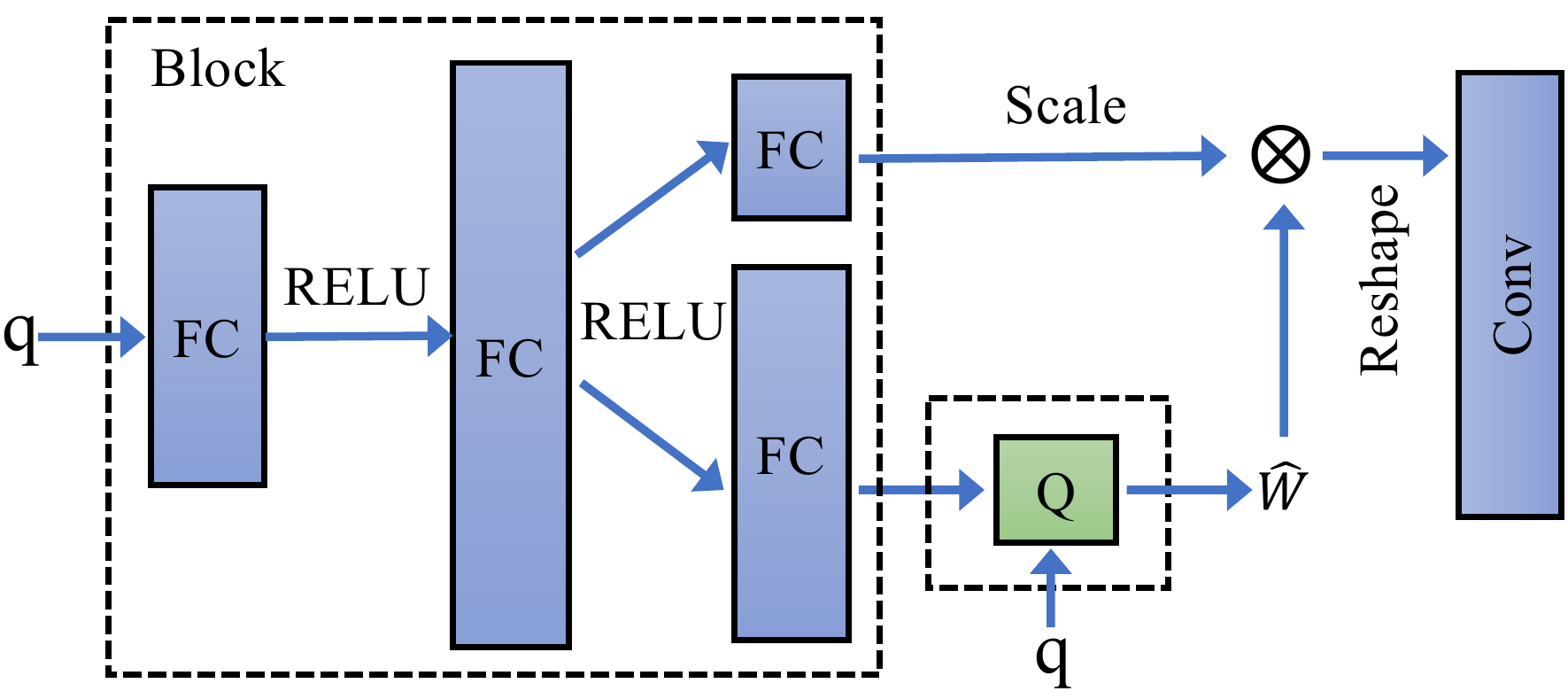}
\caption{Design of MetaQuantNet block. The block in the MetaQuantNet is a three-layered fully connected (FC) network that inputs one value q in the quantization encoding vector. The q value also controls the quantization (Q) function. We embed a branch structure after the outputs of second hidden layer. One part is input to the third hidden FC layer to obtain predicted float weights values and then quantized by the Q function as $\hat{W}$. The other part outputs a scaling parameter $\gamma$ and take $W=\gamma \hat{W}$ as the final predicted quantized weights that can be loaded to the corresponding layer (Conv or Linear layer) in the target network by reshape operation.}
\label{fig:meta_quanta_block}
\end{figure}

As shown in Figure \ref{fig:meta_quanta_block}, the MetaQuantNet block is a three-layered fully connected network (FC) with the common used activation function ReLU. The first layer takes the quantization bitwidth (q) as the input. The second hidden layer outputs are divided into two parts. One is connected to the third FC layer and Quantization function that output the weights $\hat{W}$. The other output are then connected to a third layer that outputs only one value as the scaling parameter $\gamma$. Finally, we use $W=\gamma\hat{W}$ as the quantized weights of the corresponding layer after reshape. This block network structure is quite similar to the dueling network structure with two streams in paper \cite{wang2015dueling}.

In the first training stage, the training data is input to the target network, while stochastic generated quantization vectors are input to the MetaQuantNet. The object function in Eq.(\ref{eq:loss}) is the cross-entropy loss between target network results and input ground truth. The weights of MetaQuantNet would be updated by the minibatch stochastic gradient decent (SGD) algorithms with weight decay as regularization. 

For the second search stage, since the search space is huge, we adopt the similar evolutionary algorithm as \cite{liu2019metapruning}. During search, we choose the best results that meet the constraints as parent genes to generate the off-springs. The search algorithm can be easily adapted to the our hard constraints optimization problems. Only the hybrid quantization policy that meet the constraints will be remained. In this stage, various search algorithm can also be used like Reinforcement Learning. The main difference between the RL search here and HAQ \cite{wang2019haq} is that they need retrain to obtain quantized network accuracy by finetuning float-point pretrained network while we just search the outputs of MetaQuantNet and conduct inference for target network. Hence, our method is much more efficient.

Finally in the third stage, with the searched best hybrid quantization as input, we can retrain the MetaQuantNet from scratch or just finetune to further improve the performance of target quantized network. The finetuned performance is quite good. But the retraining from scratch process can avoid local minimum in application.

%
%
%
%

\section{Experimental Results} 
In this section, we conduct extensive experiments to verify the effectiveness of the proposed method on two popular image classification datasets: CIFAR-10 and CIFAR-100 \cite{krizhevsky2014cifar}.

\subsection{Implementation Details}
The experimental target networks for quantization are the popular network structures including VGG16 \cite{simonyan2014very}, ResNet20 \cite{he2016deep} for CIFAR-10, and CIFAR-100. We use VGG16 with batch normalization and modify the structure by using one FC layer instead of the original three FC layers. Hence the total layer number are 20 and 14 for ResNet-20 and VGG16bn, respectively. VGG16bn-small model has the same structure as VGG16bn but with 4x less filter numbers than VGG16bn.

For the training stage,  we use the stochastic gradient decent (SGD) with momentum 0.9 and weight decay $10^{-4}$. The learning rate starts with 0.1 and decay half every 30 epochs after first 60 epochs. The total training process takes 200 epochs for all the full precision, quantized training and retraining stages.

\subsection{Model-Size Constrained Quantization}

\begin{table*}[htb]
\centering
\begin{tabular}{lrrrrrrrr}
\toprule
CIFAR-10&&&&ResNet20&&VGG16bn&&VGG16bn-small\\
\midrule
Scenario&Weights&Ratio&Size(MB)&Top1-Acc &Size(MB)&Top1-Acc &Size(MB)&Top1-Acc\\
\midrule
Float   		& 32bit & 1x  & 1.079 & \textbf{91.74\%}  & 58.896& \textbf{93.28\%} & 3.692& \textbf{88.91\%}\\ \midrule
Ternary  & 2bit & 16x &0.067 & 89.73\% & 3.681 &  92.33\% & 0.231 & 85.86\% \\
Binary  & 1bit & 32x &0.034 & 87.96\% & 1.840 & 91.08\% & 0.115 & 82.97\% \\ \midrule

Uniform   & 4bit    & 8x    &0.345 & 91.27\% &7.362 & 92.85\% &0.462 & 88.10\%\\
 		     	& 3bit    & 10.3x&0.105& 90.97\% & 5.718& 91.97\%  &0.358& 86.93\% \\
				& 2bit    & 16x   &0.067& 90.47\% & 3.681& 91.78\%  &0.231 & 85.62\%\\
				 & 1bit    & 32x   & 0.034&87.73\% & 1.840&90.58 \%  &0.115&80.90 \% \\ \midrule
Hybrid 	   & [1-8]bit & 10x  & 0.103 & 91.69\% &5.709 & 92.82\% &0.359 & \textbf{88.91}\% \\
				& [1-5]bit & 16x  &0.067 & 91.08\%  &3.661 & 92.47\%  &0.226 & 88.31\%\\
				& [1-5]bit & 20x   &0.0538 & 90.50\% &2.910 & 92.42\%  & 0.183 & 87.61\% \\
				 & [1-3]bit & 25x  & 0.0428& 89.72\%   & 2.355& 92.11\%  &0.147 & 86.56\% \\ 

\bottomrule
\end{tabular}
\caption{Results on CIFAR-10. We compare the top1 accuracy between uniform quantization and the searched hybrid quantization results on ResNet20, VGG16bn and VGG16bn-small networks under our proposed framework. We also give the baseline results of the traditional Binary and Ternary quantization method and Float case results. Under the same compression ratio, the hybrid quantization can always achieve higher accuracy than the uniform case.}
\label{tab:cifar10}
\end{table*}

Since we are focused to study the weights quantization of DNN, the model-size constraint for compression is studied in our experiments for simplicity. The compression ratio is defined as the ratio between float point model size and low bit quantized model size. For the N-bit uniform quantization, the model size is roughly compressed to $N/32$ and the compression ratio is $32/N$. Hence 32x compression ratio is the upper bound if using 1 bit for all the weights. We mainly focus on low bit quantization and the bitwidth are limited in [1,8] bits. In our experiments, we seek four different compression ratio: 10x, 16x, 20x and 25x. For the 10x case, we can easily generate the hybrid quantization vectors for search and training in [1,8] bits. But for higher compression ratio, we need to narrow our search space to [1,5] bits for 16x compression ratio, and [1,3] bits for even higher compression ratio.

Table \ref{tab:cifar10} shows the experimental results of CIFAR-10 for ResNet-20, VGG16bn and VGG16bn-small. We compare both uniform quantizations and hybrid quantizations under the same MetaQuantNet framework. We also give the baseline results of Binary \cite{courbariaux2015binaryconnect} and Ternary \cite{li2016ternary} quantization method and Float case results. We utilize the consistent hyperparameters for Binary and Ternary networks. For Binary network, we also introduce a learning scaling parameter $\gamma$ to scale the weights, which is different from the original work \cite{courbariaux2015binaryconnect}.

\begin{figure}[htb]
\centering
\includegraphics[width=0.45\textwidth]{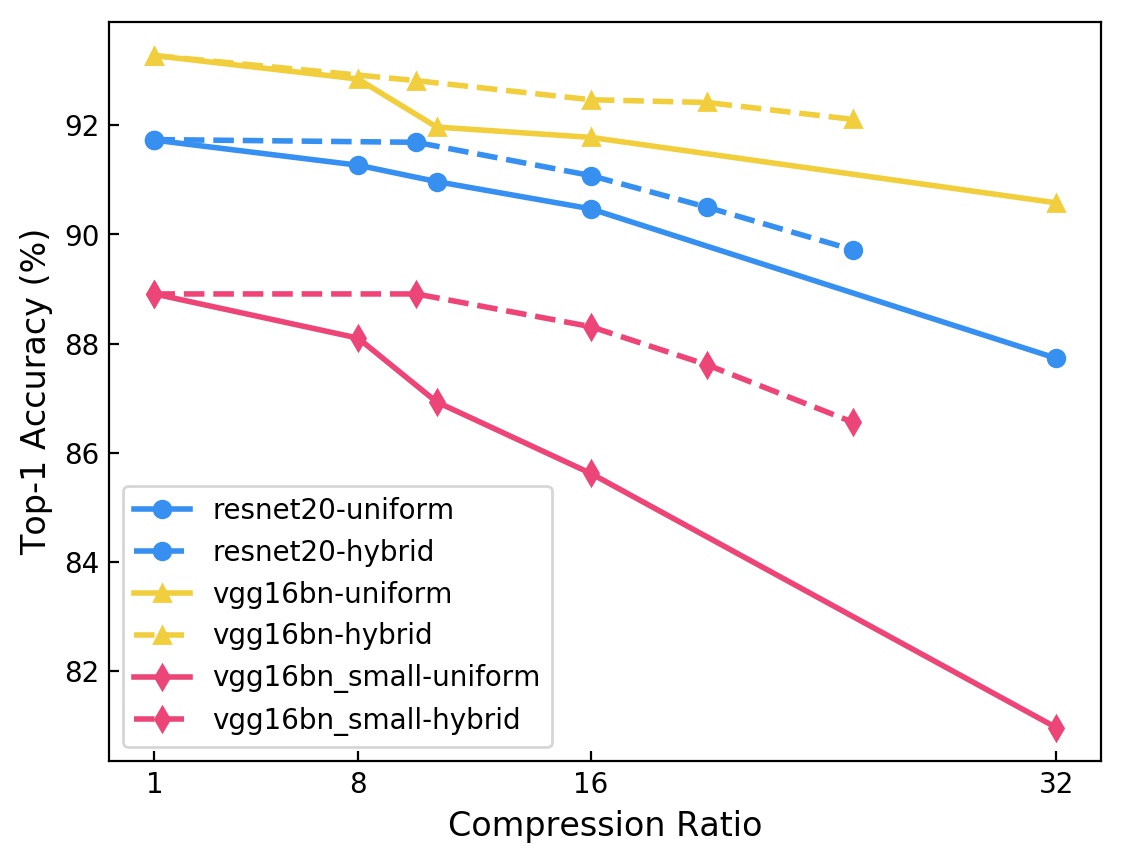}
\caption{Performance comparison between uniform quantization and hybrid quantization policy on CIFAR-10. The solid lines represent the uniform quantization and the dash lines stand for hybrid quantization case. Hybrid quantization shows better performance than uniform quantization policy.}
\label{fig:acc-vs-ratio}
\end{figure}

For clear comparison, we illustrate the performance between the uniform quantization and hybrid quantization under different compression ratio in Figure \ref{fig:acc-vs-ratio}. The solid lines represent the uniform quantization and the dash lines stand for hybrid quantization. The markers correspond to the results in Table \ref{tab:cifar10}. We can clearly observe that with the hybrid quantization policy, the quantized model accuracy drops much slower than the uniform quantization.

With increasing compression ratio, the top-1 accuracy  gradually drops for both case. But the searched hybrid quantization has much higher accuracy compared with the uniform quantization cases. Especially for a higher compression ratio, the hybrid quantization show strong capability to maintain the accuracy of compressed models. Moreover, the uniform quantization policy can just realize discrete compression ratio in 10x, 10.3x,16x and 32x. But hybrid quantizaitons can realize continuous compression ratio between 16x and 32x, in which there are still much more compression space deserve to explore. Hence only hybrid quantization can offer such abilities to achieve deeper compression in extreme low-bit quantization with higher accuracy.

Besides, for CIFAR-10 task we find that both ResNet20 and VGG16bn show strong representation capacity to maintain higher accuracy even in 1 bit quantization case. Hence, we adopt a VGG16bn-small model that behaves poorly for low bit quantization on CIFAR-10. The VGG16bn-small has four times less filters for each layer in VGG16bn. In Table \ref{tab:cifar10}, the accuracy drops significantly when using 2 bit or 1 bit uniform quantization but it can still keep better performance for hybrid quantization case even under 20x compression ratio. This trend is clearly demonstrated in Figure \ref{fig:acc-vs-ratio}.

\begin{table}[htb]
\centering
\resizebox{0.498\textwidth}{!}{
\begin{tabular}{lrrrrrr}
\toprule
CIFAR-100&&&&WRN20&&VGG16bn\\
\midrule
Scenario&Weights&Ratio&Size&Top1-Acc&Size&Top1-Acc \\ \midrule
Float   		& 32bit & 1x  & 4.342 & \textbf{72.68\%}  &59.08 & 70.63\%    \\ \midrule
Ternary & 2bit & 16x & 0.271& 69.72\% & 3.692& 70.92\%
\\
Binary & 1bit & 32x & 0.136&67.87\% & 1.846& 67.58\%
\\ \midrule
Uniform   & 4bit    & 8x  &0.543 & \textbf{71.68}\%&7.358  & 70.11\% \\ 
 			& 3bit    & 10.3x &0.422& 71.17\% &5.736& \textbf{70.94}\%   \\ 
				& 2bit    & 16x  & 0.271& 69.14\%  &3.692 & 68.91\%  \\ 
				 & 1bit    & 32x  &0.136&66.32\%  &1.846&65.10\%  \\ \midrule
Hybrid 	   & [1-8]bit & 10x  & 0.431& \textbf{71.68}\% &5.764 & 70.72\% \\
				& [1-5]bit & 16x  & 0.268 & 70.74\% & 3.686 & 70.66\% \\ 
				& [1-5]bit & 20x   &0.215 & 68.80\%  &2.893 & 70.17\%  \\ 
 				& [1-3]bit & 25x   &0.173 & 68.60\%  & 2.313& 69.05\%  \\ 

\bottomrule
\end{tabular}}
\caption{Results on CIFAR-100. We compare the top1 accuracy between uniform quantization and the searched hybrid quantization results on Wide ResNet20(WRN20) and VGG16bn networks. We also give the baseline results of Binary and Ternary  quantization method and Float case results. Under the same compression ratio, the hybrid quantization can always achieve higher accuracy than the uniform case.}
\label{tab:cifar100}
\end{table}


CIFAR-100 is a much harder task than CIFAR-10 and we use VGG16bn and Wide ResNet-20(WRN-20) as target quantized networks. The WRN-20 has the same structure as ResNet-20 but we widen the convolutional layers by adding two more feature planes, which means the widen factor $k=2$ in our experiment. Table \ref{tab:cifar100} show the results between uniform quantization and hybrid quantization strategy under various compression ratio. We also give the baseline results of Binary \cite{courbariaux2015binaryconnect} and Ternary \cite{li2016ternary} quantization method and Float case. From the results, we can obtain the same conclusion as on CIFAR-10. The hybrid quantization policy can keep a better performance accuracy under higher quantization compression ratio. The top-1 accuracy drops much slower than uniform quantization with increasing model compression.

%
%

\begin{figure}[htb]
\centering
\includegraphics[width=0.5\textwidth]{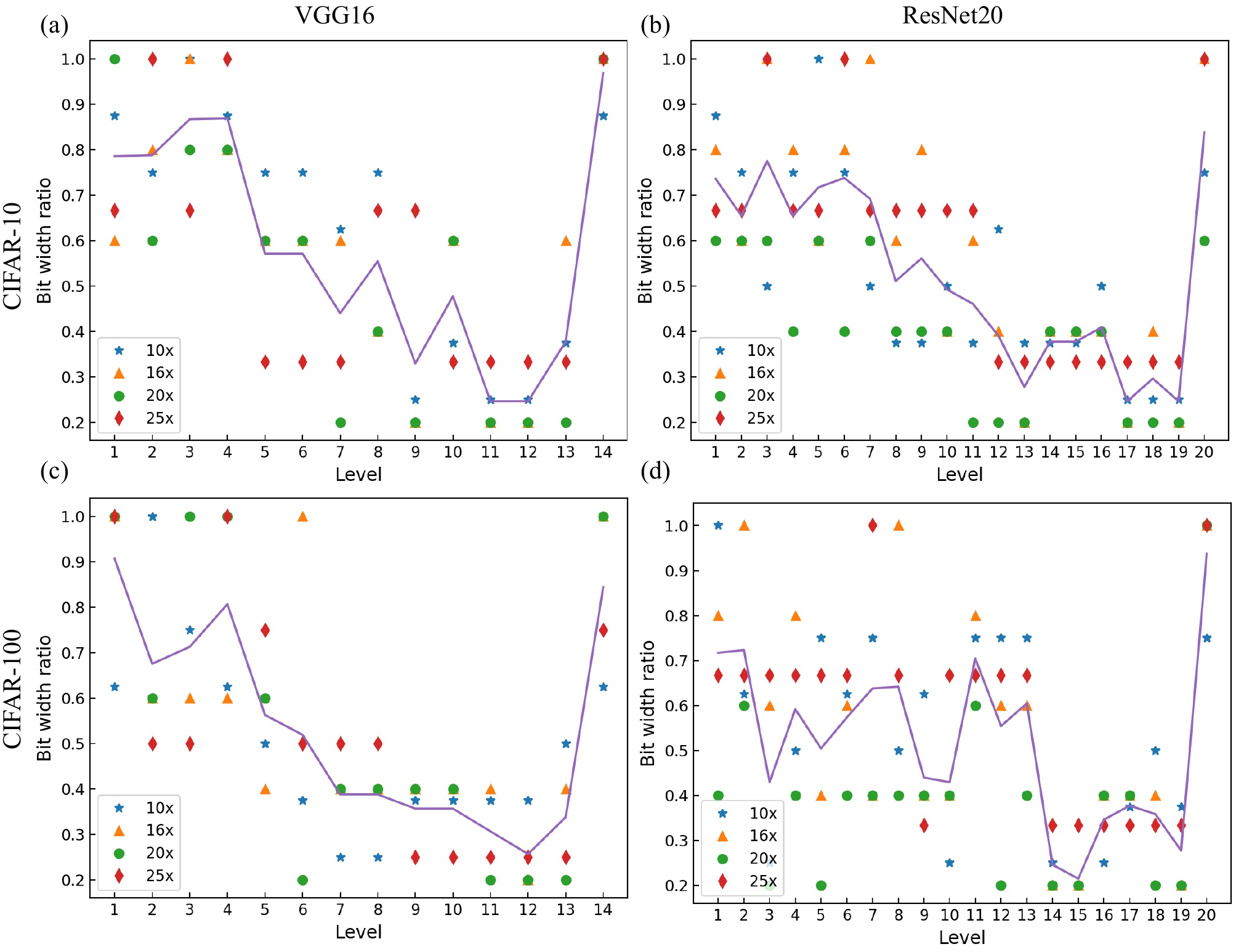}
\caption{Visualization hybrid quantization policy. (a) and (c) stand for the searched best hybrid quantization policy of VGG16bn network on CIFAR-10 and CIFAR-100. (b) and (d) stand for the searched best hybrid quantization policy of ResNet20 network on CIFAR-10 and CIFAR-100. We normalize the bitwidth as bitwidth ratio under four different compression ratio.}
\label{fig:qlevel}
\end{figure}

Furthermore, we visualization the searched best hybrid quantization policy under various compression ratio for both VGG16bn and ResNet20 network in Figure \ref{fig:qlevel}. Figure \ref{fig:qlevel}(a) and (c) stand for the searched best hybrid quantization policy of VGG16bn network on CIFAR-10 and CIFAR-100. Figure \ref{fig:qlevel}(b) and (d) stand for the searched best hybrid quantization policy of ResNet20 network on CIFAR-10 and CIFAR-100. We normalize the bitwidth of quantization encoding vector as bitwidth ratio under four different compression ratio for fair comparison. The markers represent the normalized bitwidth for each layer of target network. The solid lines are the average results accordingly. To our surprise, even though the best hybrid quantization policy varied for different task, constraints and task, the distribution of quantization encoding vector show a common pattern. The average lines show the overall trend of bitwidth importance for each layer. 

From Figure \ref{fig:qlevel}, we obtain insights that it prefers to keep higher bitwidth or precision in the first layer and last layer for both VGG16bn and ResNet20 networks no matter the task is about CIFAR-10 or CIFAR-100. This results confirm the common rule-based heuristics of quantization, such as retaining more bits in the first layer which is vital to the following layers and needs to extract low level features from raw inputs,  assigning more bitwidth to the last layer that directly computes the final outputs. Besides, the last three layers except last one layer seem not as important as other layers. They just need the lowest bitwidth representation for all the compression scenario. Such property has not been discovered before. Hence, an automatic way to realize hybrid quantization of target networks are vital for such problems.

\section{Conclusion}
In this work, we propose to use meta learning method to realize low bit hybrid quantization of neural networks automatically. The searched best hybrid quantization policy shows much better performance than the uniform quantization case.  This MetaQuantNet training, search and retraining framework inherits the advantages of meta learning  and is quite efficient and more flexible than the reinforcement learning based method. Moreover, even though the hybrid quantizaiton policy varies under different constraints, the results still show that higher bitwidth is preferred in the first layer and last classification layer, which confirms the common used design heuristics of quantization.

\bibliographystyle{named}
\bibliography{ref-hybridQ}	

\end{document}